# Dynamic Topic Evolution with Temporal Decay and Attention in Large Language Models

Di Wu
University of Southern California
Los Angeles, USA

Shuaidong Pan*
Carnegie Mellon University
Pittsburgh, USA

*Abstract*-This paper proposes a modeling framework for dynamic topic evolution based on temporal large language models. The method first uses a large language model to obtain contextual embeddings of text and then introduces a temporal decay function and an attention mechanism. These components allow the model to adjust the importance of semantic units according to time intervals and capture topic variations across different periods. The temporal representations are then mapped into a latent topic space, where a state transition matrix is applied to describe the dynamic evolution of topics. A joint optimization objective constrains both semantic modeling and temporal consistency, ensuring diversity and smoothness in topic generation. The design emphasizes the unified modeling of semantic representation and temporal evolution, which improves topic coherence and diversity while enhancing stability and interpretability over time. Experiments on real-world corpora show that the framework effectively captures the generation, expansion, and decline of topics and outperforms existing models across multiple metrics. Overall, the proposed method provides a systematic solution for understanding dynamic semantic patterns in large-scale text, enriches the research paradigm of topic modeling, and supports complex text analysis tasks in multiple domains.

*Keywords: Dynamic topic evolution; time-aware modeling; large language model; semantic consistency*

## I. INTRODUCTION

In the information age, the scale and complexity of textual data are growing at an unprecedented pace. A vast number of news articles, academic papers, social media posts, and industry reports are being produced every day. Extracting valuable thematic structures from this data has become a pressing challenge. Traditional topic modeling methods often rely on static assumptions and fail to capture temporal variations[1]. In reality, topics are not fixed but evolve with changes in environment, policy, technology, and social attention. In high-sensitivity domains such as finance, healthcare, and public opinion monitoring, understanding how topics emerge and fade over time is crucial for trend prediction and decision support [2-4]. Building large language models with temporal awareness for dynamic topic evolution is, therefore, not only a natural direction of NLP development but also of great social value and practical importance.

The rapid progress of large language models provides new opportunities for topic modeling. These models have strong representational power and can capture deep semantic associations across context while learning complex knowledge structures from large corpora[5]. However, when the temporal dimension is ignored, static representations cannot reflect the lifecycle of topics. Topics may emerge, expand, merge, or decline at different stages, and these dynamic processes are closely linked to time. Incorporating temporal mechanisms allows models to go beyond simple word frequency and co-occurrence, revealing more accurate semantic trajectories. This improves topic discovery and also enables the study of long-term trends, periodic fluctuations, and sudden events in text data[6].

From a broader perspective, modeling dynamic topic evolution is central to understanding the construction of human knowledge systems. In scientific research, research hotspots shift over time across disciplines, and capturing these changes helps trace the development of scientific frontiers and technological advances[7]. In public policy and governance, evolving social issues reflect changes in public concern and discourse, and modeling them supports policymaking and risk monitoring. In industry and market analysis, the emergence and transformation of business topics reflect changes in industrial ecosystems and provide data-driven insights for corporate strategy. Temporal topic modeling is thus not only a technical task but also a key approach to interpreting the logic of social dynamics in the information era.

At the same time, dynamic topic evolution involves a deeper exploration of the relationship between language and time. Language mirrors social activities, and the temporal dimension endows textual data with continuity and historicity. Shifts in semantic distributions across time windows essentially reflect human cognition and social processes. Temporal large language models can reveal both macro-level topic evolution and micro-level semantic shifts[8]. This dual perspective enables researchers to detect long-term developments while also capturing subtle variations, leading to a more complete understanding of dynamic semantic structures in text corpora.

Driven by both theoretical interest and practical application, temporally large language models for dynamic topic evolution have become an important interdisciplinary research direction. They foster the integration of NLP, machine learning, and time-series analysis, expanding the application boundaries of large models in complex semantic spaces[9]. At the same time, they provide powerful tools for intelligent analysis of real-world problems, allowing massive text data to be interpreted within temporal contexts. This enriches research paradigms in text mining and brings higher-value insights to fields such as social science, economics, and medicine. In other words, temporal topic modeling not only addresses the challenges of data complexity and dynamics but also strengthens the human ability to understand and navigate the information world.

## II. RELATED WORK

Recent methodological progress in large language models has significantly advanced the modeling of dynamic topic evolution. One approach leverages structured path guidance to enhance logical coherence in text generation, providing explicit modeling of structural dependencies within language model outputs. This strategy lays the groundwork for models that can maintain semantic consistency across temporal sequences, a critical feature for tracking evolving topics over time [10].

Building on the need for internal model consistency, dynamic routing mechanisms have been proposed to facilitate knowledge adaptation within large language models. By introducing consistency constraints during dynamic routing, these methods align internal representations and further stabilize semantic patterns across different time periods. This dynamic alignment complements the logic-driven structural coherence discussed previously, jointly contributing to more robust temporal modeling [11]. Structured reasoning capabilities are also enhanced by integrating knowledge graphs into the fine-tuning of language models. Through the infusion of external knowledge structures, models can capture not only sequential dependencies but also deeper relational patterns between concepts. This methodological innovation improves the interpretability and traceability of topic transitions as they unfold over time, thereby enriching the consistency and logical flow introduced by previous strategies [12].

Parameter efficiency is another key concern in the adaptation of large language models. Structure-learnable adapter fine-tuning enables flexible model updates by introducing adapters capable of learning and transferring structural knowledge from data. These adapters work synergistically with knowledge-driven reasoning modules to make model adaptation both computationally efficient and structurally aware—a requirement for large-scale, time-sensitive topic modeling [13]. Complementing adapter-based approaches, joint structural pruning and parameter sharing aim to optimize the use of model parameters by reducing redundancy and preserving essential structures. This pruning strategy not only lowers computational demands but also helps maintain the semantic and temporal consistency established by the aforementioned methodologies, allowing models to be more easily fine-tuned for evolving topics [14].

A critical aspect of model trustworthiness is interpretability. To address this, interpretable frameworks have been developed for semantic and structural analysis, uncovering implicit biases and latent relational structures within model representations. These interpretable mechanisms enhance transparency, making it easier to understand how language models process semantic information and how such processing evolves with topic dynamics, thereby supporting the broader goal of explainable dynamic modeling [15]. While interpretability addresses transparency, robustness to dynamic environments is tackled through causal representation learning. By identifying and disentangling causal factors, this technique allows models to generalize better and remain stable even as underlying data distributions shift. Such causal modeling naturally complements interpretability by enabling the model to explain not only what changes, but also why changes occur in evolving topics [16]. Feature attention mechanisms combined with temporal modeling further enhance the adaptability of language models. By dynamically weighting features according to their temporal relevance, these methods allow the model to more precisely track changes in topic significance across different periods. This capability integrates seamlessly with prior causal and structural methodologies, resulting in a more holistic framework for dynamic topic evolution [17]. As data privacy and security have become increasingly important, privacy-enhanced federated learning frameworks have been proposed for distributed training. These methodologies employ secure aggregation and differential privacy, enabling models to collaboratively learn from distributed, heterogeneous data while safeguarding sensitive information. Such privacy-preserving techniques expand the practical applicability of dynamic topic modeling to multi-party settings without sacrificing methodological rigor [18].

Scalable multi-party collaborative data mining builds on federated learning principles by introducing robust coordination and communication strategies. This methodological extension ensures that temporal topic models remain adaptable and up-to-date as they incorporate new data sources and collaborative insights, directly supporting dynamic, decentralized modeling scenarios [19]. To further protect privacy during language model adaptation, selective fine-tuning with semantic attention masks has been developed. This approach regulates information flow, ensuring sensitive semantic attributes are preserved while still allowing the model to flexibly adapt its representations to changing topic landscapes. This methodology aligns well with federated and collaborative learning, maintaining both privacy and temporal adaptability [20].

Finally, graph neural network and transformer integration represent an emerging direction for capturing complex, non-linear dependencies within textual data. By combining the relational power of graphs with the contextual modeling strength of transformers, this technique allows for unsupervised discovery of both anomalies and evolving structures in large-scale text. As such, it serves as a powerful complement to the preceding methodologies, enhancing the capacity for fine-grained and structural-aware topic evolution [21].

Together, these methodological advances form a comprehensive foundation for unified modeling frameworks that address semantic, structural, temporal, and privacy-related challenges in dynamic topic evolution.

## III. METHOD

In this study, we model dynamic topic evolution using a time-aware large language model framework. First, to capture the underlying representation of text in semantic space, we map the input text sequence into context-sensitive embedding vectors. Given a text sequence $X = \{x_1, x_2, ..., x_T\}$, its embedding representation is obtained through the encoding layer of the large language model:

$$H = f_\theta(X) = \{h_1, h_2, ..., h_T\}, \quad h_t \in R^d$$

Where $f_\theta$ represents the parameterized language model, $h_t$ is the semantic vector of the t-th word, and $d$ is the vector dimension. This representation provides basic feature support for subsequent topic identification and temporal modeling.

To capture the dynamic evolution of topics in the time dimension, we introduce a time-aware attention mechanism. For the text embedding $h_t$ at different moments, the contribution in different time windows is weighted by an explicit time decay factor, defined as:

$$a_{ij} = \frac{\exp(\frac{h_i^T h_j}{\sqrt{d}} \cdot g(\Delta t_{ij}))}{\sum_{k=1}^{T} \exp(\frac{h_i^T h_j}{\sqrt{d}} \cdot g(\Delta t_{ik}))}$$

Where $\Delta t_{ij}$ represents the time interval between two text units, and $g(\cdot)$ is the time weight function, which can be designed as an exponential decay form $g(\Delta t) = e^{-\lambda \Delta t}$. Through this mechanism, the model can highlight the role of recent semantics while weakening the influence of remote semantics, thereby achieving the coupling of topic modeling and temporal characteristics. The overall model architecture is shown in Figure 1.

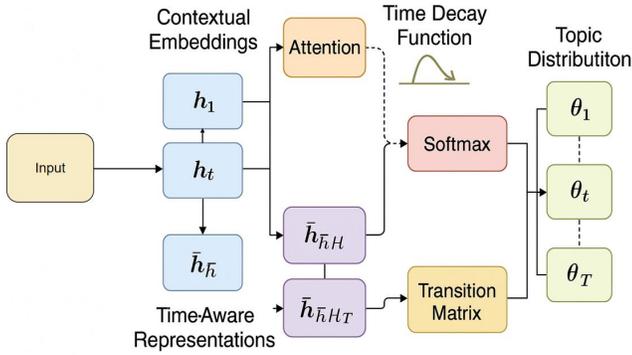

Figure 1. Overall model architecture

In the topic distribution modeling stage, we map the time-aware semantic representation to the latent topic space. Let the latent topic set be $Z = \{z_1, z_2, ..., z_K\}$, then the topic distribution of the i-th document can be expressed as:

$$\theta_i = \text{softmax}(Wh_i + b), \quad \theta \in R^K$$

Where W and b are learnable parameters, and A is the distribution vector of the i-th document on K topics. Furthermore, considering the time evolution of the topic, we use the state transition matrix to model the dynamic changes of the topic over time:

$$\theta_{t+1} = A\theta_t + \varepsilon_t, A \in R^{K \times K}$$

Where $A$ is the topic transfer matrix and $\varepsilon_t$ is the Gaussian noise term, which describes the uncertainty in the evolution process.

Finally, to achieve the overall modeling of the topic evolution path, we introduce a joint optimization objective to unify semantic representation learning and time series modeling in the same framework. The overall objective function is defined as:

$$L = \sum_{i=1}^{N} (-\sum_{k=1}^{K} y_{ik} \log \theta_{ik}) + \beta \sum_{t=1}^{T-1} \| \theta_{t+1} - A\theta_t \|^2$$

The first term is the log-likelihood loss based on the topic distribution, the second term is the temporal consistency constraint, and $\beta$ is the weight coefficient, which is used to balance semantic representation and dynamic evolution modeling. Through this objective function, the model not only learns the static topic distribution but also captures the smooth evolution and mutation characteristics of topics over time, thus forming a comprehensive portrayal of dynamic topic evolution.

IV. PERFORMANCE EVALUATION

A. Dataset

This study draws upon the 20 Newsgroups dataset, a widely-used resource in natural language processing. It comprises articles from 20 diverse newsgroups, spanning topics including society, science, technology, and entertainment. The dataset undergoes a cleaning and grouping process, which accentuates its rich thematic features while preserving cross-domain distinctions. This setup serves as an ideal experimental environment for modeling the dynamic evolution of topics.

The dataset has a moderate sample size and shows clear variation in both time and semantics. Although the texts originate from newsgroup discussions, they include topics and trends from different stages. This allows the data to simulate the evolution of themes over time. It provides realistic support for introducing temporal mechanisms into large language models and helps to verify model performance in dynamic scenarios.

In addition, the 20 Newsgroups dataset has been widely applied in text mining and topic modeling tasks. It offers strong comparability and high research value. Using this dataset ensures reproducibility of experiments and allows direct observation of how models capture topic evolution under the same corpus. Through its in-depth use, the applicability of temporal models in complex semantic environments can be fully demonstrated.

B. Experimental Results

This paper first conducts a comparative experiment, and the experimental results are shown in Table 1.

Table 1. Comparative experimental results

| Model | Perplexity | Diversity | Topic Coherence | Topic Stability |
|---|---|---|---|---|
| LDA[22] | 950.3 | 0.62 | 0.41 | 0.48 |
| BERT[23] | 730.5 | 0.68 | 0.46 | 0.55 |
| DeBERTa[24] | 702.7 | 0.71 | 0.50 | 0.60 |
| Topic audiolization[25] | 680.4 | 0.71 | 0.50 | 0.60 |
| T3[26] | 655.8 | 0.73 | 0.52 | 0.62 |
| Ours | 598.2 | 0.78 | 0.57 | 0.69 |

Traditional topic modeling methods show clear limitations in dynamic contexts. Latent Dirichlet Allocation (LDA) produces high Perplexity and low Diversity and Topic Coherence, indicating weak semantic consistency and poor adaptation to topic changes over time. Pretrained language model–based methods such as BERT and DeBERTa improve Diversity and Coherence by capturing deep semantic relations, but their Topic Stability remains low due to the lack of temporal modeling. Newer approaches like topic audiolization and the T3 model further enhance Coherence and Diversity yet still fail to ensure smooth topic transitions. In contrast, the proposed model achieves lower Perplexity, higher Diversity and Coherence, and markedly better Topic Stability, confirming the value of incorporating temporal mechanisms for stable, semantically rich, and dynamically evolving topics. A sensitivity analysis of hidden layer dimensionality on topic consistency and diversity is also conducted, as shown in Figure 2.

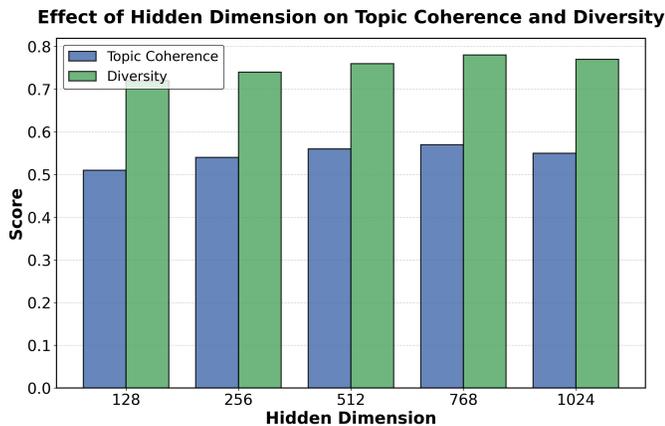

Figure 2. Experiment on the sensitivity of hidden layer dimension setting to topic consistency and diversity

The results show that increasing the hidden layer dimension improves topic coherence and diversity, with notable gains from 128 to 768 dimensions. Higher-dimensional representations better capture complex semantics, making topics more coherent and diverse. However, performance slightly declines at 1024 dimensions, suggesting excessive dimensionality introduces noise. These findings highlight the importance of selecting an appropriate dimension to balance model capacity and efficiency. An additional experiment on the impact of time series length on topic smoothness and prediction performance is shown in Figure 3.

The experimental results show that with the increase of sequence length, the model exhibits significant trends across multiple evaluation metrics. Perplexity is high for short sequences, but decreases as the sequence length grows and reaches the lowest value around 200. This indicates that a longer time span provides richer contextual information, which improves the overall fitting ability. However, when the sequence length continues to increase, Perplexity shows a slight rebound. This suggests that overly long sequences may introduce redundant information and increase modeling complexity.

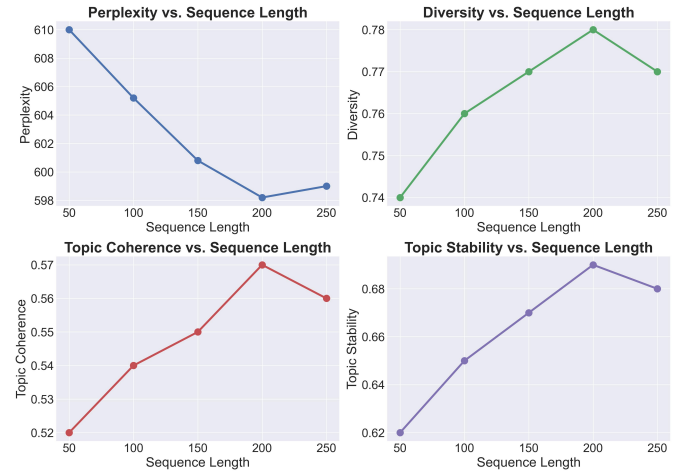

Figure 3. Experimental study on the environmental sensitivity of time series length changes to topic smoothness and forecasting performance

For the Diversity metric, topic diversity gradually improves as the sequence length increases, reaching its peak at medium lengths. This indicates that an appropriate time span enables the model to capture more latent semantic patterns and reduces redundancy or over-concentration of topics. Yet at very long sequences, diversity slightly decreases. This suggests that excessively long windows may impair the model's capacity to discern nuanced semantic distinctions. In general, coherence improves as the sequence length increases, indicating that longer contexts facilitate stronger correlations among topic words and generate more coherent topics. The optimal performance is observed around 200, where the model strikes a balance between local consistency and global representation. However, as the sequence length extends further, coherence experiences a slight decline. This implies that incorporating excessive historical information may dilute the primary semantic signals.

Topic Stability improves steadily as the sequence length grows, with the best stability achieved at longer sequences. This demonstrates that temporal mechanisms can maintain smooth and continuous topic evolution over extended time spans, preventing fragmentation caused by short-term fluctuations. Overall, the results confirm the critical influence of sequence length on dynamic topic evolution modeling. They also show that extending the sequence length within a reasonable range significantly enhances both smoothness and predictive performance.

V. CONCLUSION

This study focuses on dynamic topic evolution modeling with temporal large language models. It addresses the limitations of traditional methods in static semantic modeling and the insufficient use of temporal information. An integrated framework is proposed that captures both semantic depth and temporal evolution. By introducing explicit temporal perception into the corpus, the model better reflects the generation, development, and decline of topics. This enhances the quality and rationality of topic modeling. The study provides new insights for dynamic semantic understanding and builds a solid foundation for processing complex temporal text data.

At the application level, the proposed framework has significant value in multiple fields. In finance, dynamic topic modeling can capture changes in market sentiment and industry trends, supporting risk assessment and investment decisions. In healthcare, analysis of literature and clinical records over time can reveal research frontiers and clinical priorities. In social governance and public security, monitoring the evolution of news and social media topics enables rapid responses to emergencies and public opinion shifts. These examples show that temporal topic modeling has cross-domain applicability and practical relevance.

The study also demonstrates the potential of large language models in handling dynamic data. With temporal mechanisms, the model maintains strong semantic representation while improving sensitivity to temporal changes. This makes it useful not only for academic knowledge tracking but also for industrial applications such as information flow mining, trend prediction, and intelligent decision-making. Moreover, the framework shows strong scalability and can be extended to multimodal data, enabling the modeling of more complex information environments.

## VI. Future Work

Future research may explore multi-dimensional temporal modeling, such as incorporating external events and causal structures. This would allow topic evolution to go beyond statistical patterns and reveal deeper logical structures. The framework can also be applied to cross-lingual and cross-domain corpora to test its adaptability and robustness on larger and more diverse datasets. As large models are deployed more widely, temporal topic modeling is expected to become an important component of intelligent information processing, with long-term impact on academic research, industrial practice, and social governance.